\documentclass[runningheads]{llncs}
\usepackage{amsmath,amsfonts,amssymb}
\usepackage{graphicx}

\usepackage{physics}
\usepackage{multirow}
\usepackage[table]{xcolor}
\usepackage{comment}

\begin{document}
\title{A Comparative Evaluation of Additive Separability Tests for Physics-Informed Machine Learning}
\titlerunning{A Comparative Evaluation of Additive Separability Tests}
%
%

\author{Zi-Yu Khoo\inst{1}, Jonathan Sze Choong Low\inst{2}, and
Stéphane Bressan\inst{1}\inst{3} }
\authorrunning{Khoo et al.}

\institute{National University of Singapore. 21 Lower Kent Ridge Rd, Singapore 119077
\email{khoozy@comp.nus.edu.sg, steph@nus.edu.sg}\\
\and Singapore Institute of Manufacturing Technology, Agency for Science, Technology and Research (A*STAR), Singapore 138634 \email{sclow@simtech.a-star.edu.sg}
\and CNRS@CREATE LTD, 1 Create Way, Singapore 138602\\
}

\maketitle              
\begin{abstract}

Many functions characterising physical systems are additively separable. This is the case, for instance, of mechanical Hamiltonian functions in physics, population growth equations in biology, and consumer preference and utility functions in economics. 
We consider the scenario in which a surrogate of a function is to be tested for additive separability. The detection that the surrogate is additively separable can be leveraged to improve further learning. 
Hence, it is beneficial to have the ability to test for such separability in surrogates. The mathematical approach is to test if the mixed partial derivative of the surrogate is zero; or empirically, lower than a threshold. We present and comparatively and empirically evaluate the eight methods to compute the mixed partial derivative of a surrogate function.

\end{abstract}
\section{Introduction}
Many functions characterising physical systems are \emph{additively separable}, such as mechanical Hamiltonian functions in physics~\cite{Meyer1992}, population growth equations in biology~\cite{lotkavolterra}, and consumer preference and utility functions in economics~\cite{varian92}. 

We consider the scenario in which a machine learning model is learning a surrogate of a function without the information that it is additively separable. Testing the surrogate for the property of additive separability is of interest. Physics-informed neural networks~\cite{Karniadakis2021} combine the universal function approximation capability of neural networks~\cite{Hornik_Stinchcombe_White_1989} with the information of the symbolic properties, laws, and constraints of the underlying application domain~\cite{McDonald21,Lu22,Greydanus19,Shi_Mo_Di_2021,Zhang22}. With information regarding the additive separability of the function, physics-informed neural networks can leverage the additive separability property to improve further learning of the surrogate~\cite{khoo_22}.

The mathematical approach to test for additive separability is to check if the mixed partial derivative of the surrogate is zero. Empirically, the mixed partial derivative of the surrogate should be lower than a small threshold close to zero.

Information regarding the additive separability of a surrogate has seen several usages in physics-informed machine learning applications. In modelling Hamiltonian dynamics, for instance, Gruver et al., in~\cite{gruver2022deconstructing} argue that almost all improvement of existing works~\cite{Chen2018} is due to a second-order bias. This second-order bias results from the additive separability of the modelled Hamiltonian into the system's potential and kinetic energy as functions of position and momentum, respectively. The additive separability allowed the physics-informed neural network to "\emph{avoid[...] artificial complexity from its coordinate system}" (the input variables) and improve its performance~\cite{gruver2022deconstructing}. In symbolic regression, Udrescu and Tegmark iteratively test an unknown function for additive separability to divide a symbolic regression problem into two simpler ones that can be tackled separately. This "\emph{guarantee[s] that more accurate symbolic expressions [are] closer to the truth}", improving the performance of the symbolic regression algorithm~\cite{Udrescu_2020}. Our work to test for additive separability creates the opportunity to leverage additive separability to improve the learning of surrogates as observed in the works of Gruver and Udrescu. 

We introduce eight methods to compute the mixed partial derivative of a machine learning model, specifically a multilayer perceptron neural network learning a surrogate of a function. Our surrogate of choice is the multilayer perceptron neural network, although theoretically, any differentiable machine learning model will suffice. The first four of the eight methods compute the mixed partial derivative via finite difference of the multilayer perception neural network. Three methods arise from the different methods to automatically compute mixed partial derivatives using automatic differentiation of the multilayer perceptron neural network. The last method arises from the symbolic differentiation of a multilayer perceptron neural network. We present and comparatively and empirically evaluate the performance of eight methods in computing the mixed partial derivative of the surrogate functions.

The remainder of this paper is structured as follows. Section~\ref{sec:background} presents the necessary background for additive separability and multilayer perception neural network surrogates. Section~\ref{sec:methodology} presents the eight methods to compute the mixed partial derivative of a multilayer perception neural network. Section~\ref{sec:exp} presents and discusses the results of an empirical comparative evaluation of the eight methods. Section~\ref{sec:conclusion} concludes the paper. 

\section{Background and Related Work} \label{sec:background}
\subsection{Additive Separability}
An additively separable real function $f(\vec{x},\vec{y}) \in \mathbb{R}$ is of the form $f(\vec{x},\vec{y}) = g(\vec{x}) + h(\vec{y})$, where $\vec{x} \in \mathbb{R}^n$, $\vec{y}\in \mathbb{R}^m$ are vectors representing disjoint subsets of $\mathbb{R}^{m+n}$ input variables and $g(\vec{x}), f(\vec{y}) \in \mathbb{R}$. $x_n$ and $y_m$ denote elements of $\vec{x}$ and $\vec{y}$.

The necessary and sufficient condition for a function to be additively separable is that the mixed partial derivative of the function equals zero. The mixed partial derivative is the second derivative of the function. The first derivative is taken with respect to an element in either $\vec{x}$ or $\vec{y}$, and the second derivative is taken with respect to an element in either $\vec{y}$ or $\vec{x}$. This condition is shown in Equation~\ref{eqn:addsep_hamiltonian_mixedpartial}, where the mixed partial derivative is found with respect to $x_n$, an element in $\vec{x}$ first, then $y_m$, an element in $\vec{y}$.
\begin{align}
    \pdv{f(\vec{x},\vec{y})}{x_n}{y_m} = \pdv{}{y_m}\left( \pdv{(g(\vec{x})+h(\vec{y}))}{x_n} \right) =  \pdv{}{y_m} \left( \pdv{g(\vec{x})}{x_n}\right) = 0 \label{eqn:addsep_hamiltonian_mixedpartial}
\end{align}

Finite difference methods refer to those that obtain a numerical solution for partial derivatives by replacing the derivatives with their appropriate numerical differentiation formulae. In general, a finite difference approximation of the value of some derivative of a scalar function $f(x)$ at a point in its domain relies on a suitable combination of sampled function values at its nearby points~\cite{olver2013pde}. 

Starting with the first-order derivative, the simplest finite difference approximation for a multivariate function $f(\vec{x},\vec{y})$ is the ordinary difference quotient shown in Equation~\ref{eqn:finite_difference_first} where the function $f(\cdot)$ is sampled at $f(\vec{x}, \vec{y})|_{x_n=x_N+h}$ and $f(\vec{x}, \vec{y})|_{x_n=x_N}$ and all other elements are kept constant. $x_N$ is a scalar sample of the element $x_n$ in $\vec{x}$, and $h$ is the scalar distance between the two samples of $x_n$. Indeed, if $f(\cdot)$ is differentiable at $x_N$ then $\pdv{f(\vec{x},\vec{y})}{x}|_{x = x_N}$ is by definition, the limit, as $h \rightarrow 0$ of the finite difference quotients~\cite{olver2013pde}.
\begin{equation}
    \pdv{f(\vec{x},\vec{y})}{\vec{x}}|_{x_n = x_N} = \frac{f(\vec{x}, \vec{y})|_{x_n = x_N+h} - f(\vec{x}, \vec{y})|_{x_n=x_N}}{h} \label{eqn:finite_difference_first}
\end{equation}

The finite difference for a multivariate function is analogous to partial derivatives in several variables. The finite difference analogue of Equation~\ref{eqn:addsep_hamiltonian_mixedpartial} is shown in Equation \ref{eqn:finite_difference_second}. The components of the numerator of Equation~\ref{eqn:finite_difference_second} are defined in Equations~\ref{eqn:def1} to \ref{eqn:def4}. $x_N$ and $y_M$ are scalar samples of elements $x_n$ and $y_m$ respectively, and $h$ and $k$ are scalar. All other elements are kept constant.
\begin{align} 
    &\pdv{f(\vec{x},\vec{y})}{x_n}{y_m} \bigg\rvert_{\substack{x_n = x_{N}\\ y_m = y_{M}}} \notag \\
    &= \frac{f(x_{N}+h, y_{M}+k) - f(x_{N}+h, y_{M}) - f(x_{N}, y_{M}+k) + f(x_{N}, y_{M})}{h\times k} \label{eqn:finite_difference_second}
\end{align}
\begin{align}
    f(x_{N}+h, y_{M}+k) &= f(\vec{x},\vec{y})|_{x_n = x_{N}+h, y_m = y_{M}+k} \label{eqn:def1} \\
    f(x_{N}+h, y_{M}) &= f(\vec{x},\vec{y})|_{x_n = x_{N}+h, y_m = y_{M}} \label{eqn:def2} \\
    f(x_{N}, y_{M}+k) &= f(\vec{x},\vec{y})|_{x_n = x_{N}, y_m = y_{M}+k} \label{eqn:def3} \\
    f(x_{N}, y_{M}) &= f(\vec{x},\vec{y})|_{x_n = x_{N}, y_m = y_{M}} \label{eqn:def4} 
\end{align}

Seminal works that test for additive separability of a function make use of the finite difference of the function. Udrescu et al.~\cite{Udrescu_2020} and Bellenot~\cite{bellenot_addsep} state that for a function to be additively separable, for every pair of samples $({x}_a,{y}_a)$ and $({x}_b,{y}_b)$  where $x$ is to be additively separable from $y$, the difference between the two pairwise sums of the values of the function at diagonally opposite corners of the rectangle $\langle({x}_a,{y}_a), ({x}_b,{y}_a), ({x}_b,{y}_b), ({x}_a,{y}_b)\rangle$ equals zero~\cite{bellenot_addsep,Udrescu_2020}. This is shown for a bivariate function $f(x,y)$ in Equation~\ref{eqn:addsep_diagoppori}. 
\begin{align} 
    \forall {x_a}, {x_b} \in x &\quad  \forall {y_a}, {y_b} \in y \notag \\
    &\left(f({x}_a,{y}_a) + f({x}_b,{y}_b)\right) - \left(f({x}_a,{y}_b) + f({x}_b,{y}_a) \right) = 0 \label{eqn:addsep_diagoppori}
\end{align}
Equation~\ref{eqn:addsep_diagoppori} and Equation~\ref{eqn:finite_difference_second} are equivalent in the case where Equation~\ref{eqn:finite_difference_second} describes a bivariate function with the inputs $x_N$ and $y_M$, after substituting $x_N = x_a$, $x_N+h = x_b$, $y_M = y_a$, $y_M+k = y_b$, and $h=k=1$.  

Equation~\ref{eqn:addsep_diagoppori} can be generalised to test multivariate functions for additive separability. A multivariate function $f(\vec{x}, \vec{y})$ is additive separable if Equation~\ref{eqn:addsep_diagoppori} holds, and all elements of $\vec{x}$ and $\vec{y}$ except $x_N$ and $y_M$ are kept constant.



The set of additively separable functions is closed under operations including addition, multiplication by constants, partial derivatives, and integrals with respect to either of the disjoint subsets of input variables~\cite{bellenot_addsep}. Therefore a test for additive separability only has to decompose a function into two components and can be applied repeatedly to a function that is multiply additively separable. 


Several other tests for additive separability have been proposed in the literature, in particular in economics. These tests make assumptions that limit their applicability to specific families of functions. The seminal work of Leontief~\cite{leontief_1947} proposed a test for additive separability for functions with three or more variables and non-zero first derivatives. Gorman~\cite{gorman_68}, Varian~\cite{varian83}, Diewert and Parkan~\cite{DIEWERT1985127} and Fleissig and Whitney~\cite{FLEISSIG2007215} developed tests for additive separability that were specific to concave and monotonic functions. Polisson, Quah and Renou~\cite{polisson20} and Polisson~\cite{polisson18} developed tests for additive separability that assume non-satiation or a positive correlation between the input variables and output of a function.

\subsection{Multilayer Perceptron Neural Network Surrogates}
Multilayer perceptrons are regression tools~\cite{hastie08} that are universal function approximators~\cite{Hornik_Stinchcombe_White_1989}. They can approximate any function to any degree of accuracy from one finite dimensional space to another. 

We consider a multilayer perceptron neural network with one input layer, one hidden layer, and one output layer. The multilayer perception neural network regresses an output, shown in Equation~\ref{eqn:method8:output}. In Equation~\ref{eqn:method8:output}, the neural network has two inputs, $x_1$ and $x_2$, and one output, $f(x_1,x_2)$. $\sigma_1$ and $\sigma_2$ are the activation functions for the input and hidden layer respectively. $F^\intercal$ is the 3 by 2 weight matrix of the input layer, $G^\intercal$ is the 3 by 3 weight matrix of the hidden layer, and $H^\intercal$ is the 1 by 3 weight matrix of the output layer. $B$ is the 1 by 3 bias matrix of the input layer, and $C$ is the 1 by 3 bias matrix of the hidden layer. The elements of $F^\intercal$, $G^\intercal$, $H^\intercal$, $B$ and $C$ are denoted by $w$, $v$, $n$, $b$ and $c$ respectively. Their subscripts indicate the row and column of the element in the matrix. The output of the input layer is denoted as $W$. 

\begin{align}
    f(x_1,x_2)  &= n_1 \sigma_2 (v_{11}  \sigma_1(W_1) + v_{12} \sigma_1(W_2) + v_{13} \sigma_1(W_3)+c_1)  \notag \\
     &+n_2 \sigma_2 (v_{21} \sigma_1(W_1) + v_{22} \sigma_1(W_2) + v_{23} \sigma_1(W_3) +c_2) \notag  \\
     &+n_3 \sigma_2 (v_{31} \sigma_1(W_1) + v_{32} \sigma_1(W_2) + v_{33} \sigma_1(W_3) +c_3) \label{eqn:method8:output} 
\end{align}
Where the outputs of the input layer are denoted as $W$ and their subscripts enumerate the three outputs of the layer.
\begin{align}
     W_1 &= w_{11}x_1 +w_{12}x_2 +b_1 \notag \\
     W_2 &= w_{21}x_1 +w_{22}x_2 +b_2 \notag \\
     W_3 &= w_{31}x_1 +w_{32}x_2 +b_3 
\end{align}

In the context of additively separable functions and their mixed partial derivatives, the derivatives of the output of the multilayer perceptron neural network should be considered. The first derivative of the output of the multilayer perceptron neural network, $f(x_1,x_2)$, with respect to its input, $x_1$, is computed in Equation~\ref{eqn:first_derivative_symbolic}. $\sigma_{1,x_1}^\prime$ and $\sigma_{2,x_1}^\prime$ is the first derivative of the activation functions $\sigma_1$ and $\sigma_2$ with respect to $x_1$. The output of the hidden layer is denoted as $V$.
\begin{align}
    & \pdv{f(x_1,x_2)}{x_1} \notag \\
    & \quad = n_1 \sigma_{2,x_1}^\prime (V_1) \times (v_{11}w_{11}\sigma_{1,x_1}^\prime (W_1) + v_{12}w_{21}\sigma_{1,x_1}^\prime (W_2) + v_{13}w_{31}\sigma_{1,x_1}^\prime (W_3)) \notag \\
    & \quad + n_2 \sigma_{2,x_1}^\prime (V_2) \times (v_{21}w_{11}\sigma_{1,x_1}^\prime (W_1) + v_{22}w_{21}\sigma_{1,x_1}^\prime (W_2) + v_{23}w_{31}\sigma_{1,x_1}^\prime (W_3)) \notag \\
    & \quad + n_3 \sigma_{2,x_1}^\prime (V_3) \times (v_{31}w_{11}\sigma_{1,x_1}^\prime (W_1) + v_{32}w_{21}\sigma_{1,x_1}^\prime (W_2) + v_{33}w_{31}\sigma_{1,x_1}^\prime (W_3)) \label{eqn:first_derivative_symbolic}
\end{align}
Where 
\begin{align}
    V_1 &= v_{11} \sigma_1 (W_1)+v_{12} \sigma_1 (W_2)+v_{13} \sigma_1 (W_3) +c_1 \notag \\
    V_2 &= v_{21} \sigma_1 (W_1)+v_{22} \sigma_1 (W_2)+v_{23} \sigma_1 (W_3) +c_2 \notag \\
    V_3 &= v_{31} \sigma_1 (W_1)+v_{32} \sigma_1 (W_2)+v_{33} \sigma_1 (W_3) +c_3 \notag 
\end{align}

The mixed partial derivative of the multilayer perceptron neural network is shown in Equation~\ref{eqn:method8:deriv} where $\sigma_{1,x_1}^\prime$, $\sigma_{1,x_2}^\prime$, $\sigma_{2,x_1}^\prime$ and $\sigma_{2,x_2}^\prime$ are the first derivatives for the activation functions for the input and hidden layer with respect to $x_1$ and $x_2$ respectively, and $\sigma_1^{\prime\prime}$ and $\sigma_2^{\prime\prime}$ are the second derivatives for the activation functions for the input and hidden layer with respect to both $x_1$ and $x_2$ respectively. 

\begin{align}
    &\pdv{f(x_1,x_2)}{x_1}{x_2} \notag \\
    & \quad= n_1 \sigma_2^{\prime\prime} (V_1) \times (VW_{1,x_1}) \times (VW_{1,x_2}) +n_1 \sigma_{2,x_1}^\prime (V_1) \times (VW_1^{\prime\prime})  \notag \\
    & \quad= n_2 \sigma_2^{\prime\prime} (V_2) \times (VW_{2,x_1}) \times (VW_{2,x_2}) +n_2 \sigma_{2,x_1}^\prime (V_2) \times (VW_2^{\prime\prime})  \notag \\
    & \quad= n_3 \sigma_2^{\prime\prime} (V_3) \times (VW_{3,x_1}) \times (VW_{3,x_2}) +n_3 \sigma_{2,x_1}^\prime (V_3) \times (VW_3^{\prime\prime})     \label{eqn:method8:deriv}
\end{align}
Where $VW_{1,x_1}^\prime$, $VW_{1,x_2}^\prime$, $VW_{2,x_1}^\prime$, $VW_{2,x_2}^\prime$, $VW_{3,x_1}^\prime$, $VW_{3,x_2}^\prime$, $VW_1^{\prime\prime}$, $VW_2^{\prime\prime}$ and $VW_3^{\prime\prime}$ are computed using the chain rule. 


\begin{align}
    VW_{1,x_1}^\prime &= v_{11}\times w_{11} \times \sigma_{1,x_1}^\prime(W_1) + v_{12}\times w_{21} \times \sigma_{1,x_1}^\prime(W_2) + v_{13} \times w_{31}\times \sigma_{1,x_1}^\prime(W_3) \notag \\
    VW_{1,x_2}^\prime &= v_{11}\times w_{21} \times \sigma_{1,x_2}^\prime(W_1) + v_{12}\times w_{22} \times \sigma_{1,x_2}^\prime(W_2) + v_{13} \times w_{32}\times \sigma_{1,x_2}^\prime(W_3) \notag \\
    VW_1^{\prime\prime} &= v_{11} \times w_{11} \times w_{12}  \times \sigma_1^{\prime\prime} (W_1) + v_{12} \times w_{21} \times w_{22} \times \sigma_1^{\prime\prime} (W_2) \notag \\
    & \quad +v_{13} \times w_{31} \times w_{32} \times \sigma_1^{\prime\prime} (W_3) \notag 
\end{align}
\begin{align}
    VW_{2,x_1}^\prime &= v_{21}\times w_{11} \times \sigma_{1,x_1}^\prime(W_1) + v_{22}\times w_{21} \times \sigma_{1,x_1}^\prime(W_2) + v_{23} \times w_{31}\times \sigma_{1,x_1}^\prime(W_3) \notag \\
    VW_{2,x_2}^\prime &= v_{21}\times w_{21} \times \sigma_{1,x_2}^\prime(W_1) + v_{22}\times w_{22} \times \sigma_{1,x_2}^\prime(W_2) + v_{23} \times w_{32}\times \sigma_{1,x_2}^\prime(W_3) \notag \\
    VW_2^{\prime\prime} &= v_{21} \times w_{11} \times w_{12}  \times \sigma_1^{\prime\prime} (W_1) + v_{22} \times w_{21} \times w_{22} \times \sigma_1^{\prime\prime} (W_2) \notag \\
    & \quad +v_{23} \times w_{31} \times w_{32} \times \sigma_1^{\prime\prime} (W_3) \notag 
\end{align}
\begin{align}
    VW_{3,x_1}^\prime &= v_{31}\times w_{11} \times \sigma_{1,x_1}^\prime(W_1) + v_{32}\times w_{21} \times \sigma_{1,x_1}^\prime(W_2) + v_{33} \times w_{31}\times \sigma_{1,x_1}^\prime(W_3) \notag \\
    VW_{3,x_2}^\prime &= v_{31}\times w_{21} \times \sigma_{1,x_2}^\prime(W_1) + v_{32}\times w_{22} \times \sigma_{1,x_2}^\prime(W_2) + v_{33} \times w_{32}\times \sigma_{1,x_2}^\prime(W_3) \notag \\
    VW_3^{\prime\prime} &= v_{31} \times w_{11} \times w_{12}  \times \sigma_1^{\prime\prime} (W_1) + v_{32} \times w_{21} \times w_{22} \times \sigma_1^{\prime\prime} (W_2) \notag \\
    & \quad +v_{33} \times w_{31} \times w_{32} \times \sigma_1^{\prime\prime} (W_3) \notag 
\end{align}

\section{Methodology} \label{sec:methodology}

We consider the scenario in which a machine learning model is a surrogate of a function without the information regarding additive separability of the function. The surrogate is a multilayer perceptron neural network that learns a multivariate function $f(\vec{x},\vec{y})$. We design eight methods to compute the mixed partial derivative of the surrogate, to test the surrogate, and hence the unknown function, for additive separability. This section describes the design of the eight methods.

The first four methods compute the mixed partial derivative via finite difference. Three methods arise from the different methods to automatically compute mixed partial derivatives using automatic differentiation of the surrogate. The last method arises from the symbolic differentiation of the surrogate.

Method 1 computes the mixed partial derivative via Equation~\ref{eqn:finite_difference_second} by evaluating the surrogate at $(x_{N}, y_{M})$, $(x_{N}+h, y_{M})$, $(x_{N}, y_{M}+k)$ and $(x_{N}+h, y_{M}+k)$ in Equation~\ref{eqn:finite_difference_second}, and setting $h=k=1$. 

Method 2 computes the mixed partial derivative via Equation~\ref{eqn:finite_difference_second} by evaluating the surrogate at $(x_{N}, y_{M})$, $(x_{N}+h, y_{M})$, $(x_{N}, y_{M}+k)$ and $(x_{N}+h, y_{M}+k)$ in Equation~\ref{eqn:finite_difference_second}, and setting $h$ and $k$ to be the distances between $x_{N}$ and $x_{N}+h$, $y_{M}$ and $y_{M}+k$ respectively. 

Method 3 computes the mixed partial derivative via Equation~\ref{eqn:finite_difference_second} by evaluating the surrogate at $(x_{N}, y_{M})$, $(x_{N}+h, y_{M})$, $(x_{N}, y_{M}+k)$ and $(x_{N}+h, y_{M}+k)$ in Equation~\ref{eqn:finite_difference_second}. However, it defines $x_{N}+h$ and $y_M+k$ to be the median of $x_{N}$ and $y_{M}$ respectively. It sets $h=k=1$. We note that this is the methodology used by Udrescu et al. in their symbolic regression algorithm, AI Feynman~\cite{Udrescu_2020}.

Method 4 computes the mixed partial derivative via Equation~\ref{eqn:finite_difference_second} by evaluating the surrogate at $(x_{N}, y_{M})$, $(x_{N}+h, y_{M})$, $(x_{N}, y_{M}+k)$ and $(x_{N}+h, y_{M}+k)$ in Equation~\ref{eqn:finite_difference_second}. However, it defines $x_{N}+h$ and $y_M+k$ to be the median of $x_{N}$ and $y_{M}$ respectively. It sets $h$ and $k$ to be the distances between $x_{N}$ and $x_{N}+h$, $y_{M}$ and $y_{M}+k$ respectively. 

We note that Methods 1 and 2 require a quadratic number of evaluations of the surrogate, while Methods 3 and 4 require a linear number of evaluations of the surrogate.

Method 5 computes the mixed partial derivative via Equation~\ref{eqn:addsep_hamiltonian_mixedpartial}. The mixed partial derivative of the surrogate is computed using automatic differentiation, by taking the first derivative of the surrogate with respect to an element in $\vec{x}$, then taking a second derivative of the surrogate with respect to an element in $\vec{y}$.

Method 6 computes the mixed partial derivative via Equation~\ref{eqn:addsep_hamiltonian_mixedpartial}. The mixed partial derivative of the surrogate is computed using automatic differentiation, by taking the first derivative of the surrogate with respect to an element in $\vec{y}$, then taking a second derivative of the surrogate with respect to an element in $\vec{x}$.

Method 7 computes the mixed partial derivative via Equation~\ref{eqn:addsep_hamiltonian_mixedpartial}. The mixed partial derivative of the surrogate is computed using automatic differentiation, by finding the Hessian of the surrogate with respect to an element in $\vec{x}$ and an element in $\vec{y}$. 

Method 8 computes the mixed partial derivative of a surrogate multilayer perceptron neural network symbolically, following Equation~\ref{eqn:method8:deriv}. Given a surrogate of a function, it creates a second surrogate multilayer perceptron neural network with the same weights and biases. This second surrogate multilayer perception neural network instead models the mixed partial derivative of the unknown function. The new surrogate multilayer perception neural network has layers and activations following Equation~\ref{eqn:method8:deriv}. The inputs of the second surrogate are the same as the first. The mixed partial derivative of the unknown function is the output of the new surrogate multilayer perceptron neural network.

\section{Performance Evaluation} \label{sec:exp}
Eight classifiers are created based on the eight methods listed in Section~\ref{sec:methodology}. Independently, functions that are either additively or non-additively separable are created, and one surrogate is trained on each function. Each classifier is then given a trained surrogate. The eight classifiers each return a test output. The test outputs are aggregated to compare the eight classifiers. This section presents and discusses the results of the comparison of the eight methods. 

\subsection{Experimental Setup}

\subsubsection{Setup of the Additively and Non-Additively Separable Surrogates}
We create unknown functions that are either additively or non-additively separable. We train one surrogate for each unknown function. 


We create two- and three-variabled unknown functions comprising additive and multiplicative combinations of polynomial, trigonometric, exponential, radical and logarithmic uni-variate sub-functions, shown in Table~\ref{tab:subfunctions}. A total of $3744$ additively and non-additively separable unknown functions were created.
\begin{table}[]
    \caption{Twelve sub-functions with input $n$, a placeholder for variables $x$, $y$ and $z$}
    \centering \scriptsize
    \setlength{\tabcolsep}{12pt}
    \begin{tabular}{|c|c|c|c|}
    \hline
    \multicolumn{4}{|c|}{Sub-functions} \\
    \hline
    $f(n) = n$ & $f(n) = n^2$  & $f(n) = (\frac{n}{3})^3$ &$f(n) = \frac{1}{n+4}$ \\
    $f(n) = \sin(n)$ & $f(n) = \cos(n)$ & $f(n) = \sin(n)^2$ & $f(n) = \cos(n)^2$ \\
    $f(n) = \exp(n)$ & $f(n) = \log(n+4)$ & $f(n) = \sqrt{|n|}$ & $f(n) = n^{1/3}$  \\
    \hline
    \end{tabular}
    \label{tab:subfunctions}
\end{table}

One multilayer perceptron neural network surrogate is trained per unknown function. We select $30$ data points uniformly at random within the range of $[-3,3]$ for each input variable of each unknown function\footnote{We note that if data is generated evenly in a grid, Equation~\ref{eqn:addsep_hamiltonian_mixedpartial} (after computing the derivative between consecutive points) and Equation~\ref{eqn:addsep_diagoppori} can be evaluated for each unknown function immediately without a surrogate.}. The data is input to the analytical form of the created unknown functions to get function outputs, which we call output data. Tuples of the input and output data are used as training data for a surrogate multilayer perceptron. Each surrogate multilayer perception neural network has two hidden layers of width 26 with softplus activation, mean squared error loss, batch size of $128$ and Adam optimizer with learning rate $0.01$. $80\%$ of the data is used for training and $20\%$ for validation. All models are trained to convergence using a validation-based dynamic stopping criteria~\cite{Prechelt1998} with patience of $500$ epochs. 
All models are trained and evaluated on two GeForce GTX1080 GPUs, with 64 GB of RAM and 12 processors.

\subsubsection{Setup of the Eight Classifiers}
The eight classifiers make use of the trained surrogate for each unknown function, $\hat{f}(\vec{x},\vec{y})$ to evaluate additive separability. All eight classifiers compute the mixed partial derivative of the surrogate with respect to $x_1$ and $y_1$, which are the first elements in $\vec{x}$ and $\vec{y}$ respectively. The values of all other inputs to the surrogates are kept constant for each evaluation of each classifier. The eight classifiers compute the mixed partial derivatives on a test dataset, comprising $30$ data points generated uniformly in a grid within the range of $[-3,3]$ for each input variable of each unknown function.

Classifier 1 computes the mixed partial derivative via Equation~\ref{eqn:finite_difference_second} by evaluating the surrogate at $\hat{f}(x_{1}, y_{1})$, $\hat{f}(x_{1}+h, y_{1})$, $\hat{f}(x_{1}, y_{1}+k)$ and $\hat{f}(x_{1}+h, y_{1}+k)$. The pair of points $(x_{1}, y_{1})$ and $(x_{1}+h, y_{1}+k)$ correspond to all pairwise combinations of samples from the test dataset. $h$ and $k$ are set to 1. For a test dataset comprising $n$ tuples of $(\vec{x}, \vec{y})$, this corresponds to $\frac{n(n-1)}{2}$ combinations of samples. The mixed partial derivative is averaged over all $\frac{n(n-1)}{2}$ combinations of samples for a comparative evaluation. 

Classifier 2 computes the mixed partial derivative via Equation~\ref{eqn:finite_difference_second} by evaluating the surrogate at $\hat{f}(x_{1}, y_{1})$, $\hat{f}(x_{1}+h, y_{1})$, $\hat{f}(x_{1}, y_{1}+k)$ and $\hat{f}(x_{1}+h, y_{1}+k)$. The pair of points $(x_{1}, y_{1})$ and $(x_{1}+h, y_{1}+k)$ correspond to all pairwise combinations of samples from the test dataset. $h$ and $k$ are set to be the distances between each pair of samples of $(x_{1}, y_{1})$ and $(x_{1}+h, y_{1}+k)$. For a test dataset comprising $n$ tuples of $(\vec{x}, \vec{y})$, this corresponds to $\frac{n(n-1)}{2}$ combinations of samples. The mixed partial derivative is averaged over all $\frac{n(n-1)}{2}$ combinations of samples for a comparative evaluation. 

Classifier 3 computes the mixed partial derivative via Equation~\ref{eqn:finite_difference_second} by evaluating the surrogate at $\hat{f}(x_{1}, y_{1})$, $\hat{f}(x_{1}+h, y_{1})$, $\hat{f}(x_{1}, y_{1}+k)$ and $\hat{f}(x_{1}+h, y_{1}+k)$.  $x_{1}+h$ and $y_{1}+k$ are the median from all samples of $x_{1}$ and $y_{1}$ from the test dataset respectively. $h$ and $k$ are set to 1. For a test dataset comprising $n$ tuples of $(\vec{x}, \vec{y})$, the mixed partial derivative is averaged over all $n$ samples for a comparative evaluation.

Classifier 4 computes the mixed partial derivative via Equation~\ref{eqn:finite_difference_second} by evaluating the surrogate at $\hat{f}(x_{1}, y_{1})$, $\hat{f}(x_{1}+h, y_{1})$, $\hat{f}(x_{1}, y_{1}+k)$ and $\hat{f}(x_{1}+h, y_{1}+k)$.  $x_{1}+h$ and $y_{1}+k$ are the median from all samples of $x_{1}$ and $y_{1}$ from the test dataset respectively. $h$ and $k$ are set to be the distances between each pair of samples of $(x_{1}, y_{1})$ and $(x_{1}+h, y_{1}+k)$. For a test dataset comprising $n$ tuples of $(\vec{x}, \vec{y})$, the mixed partial derivative is averaged over all $n$ samples for a comparative evaluation. 

Classifier 5 computes the mixed partial derivative via Equation~\ref{eqn:addsep_hamiltonian_mixedpartial}. The mixed partial derivative of the surrogate, $\hat{f}(x_{1}, y_{1})$, is computed using automatic differentiation via \verb|autograd| in \verb|pytorch|~\cite{pytorch-autodiff,NEURIPS2019_9015}, at all samples $(x_{1}, y_{1})$ in the test dataset. This is done by taking the first derivative of the surrogate with respect to $x_1$, then finding the derivative of the surrogate again, but with respect to $y_1$. For a test dataset comprising $n$ tuples of $(\vec{x}, \vec{y})$, the mixed partial derivative is averaged over all $n$ samples for a comparative evaluation. 

Classifier 6 computes the mixed partial derivative via Equation~\ref{eqn:addsep_hamiltonian_mixedpartial}. The mixed partial derivative of the surrogate, $\hat{f}(x_{1}, y_{1})$, is computed using automatic differentiation via \verb|autograd| in \verb|pytorch|~\cite{pytorch-autodiff,NEURIPS2019_9015}, at all samples $(x_{1}, y_{1})$ in the test dataset. This is done by taking the first derivative of the surrogate with respect to $y_1$, then finding the derivative of the surrogate again, but with respect to $x_1$. For a test dataset comprising $n$ tuples of $(\vec{x}, \vec{y})$, the mixed partial derivative is averaged over all $n$ samples for a comparative evaluation. 

Classifier 7 computes the mixed partial derivative via Equation~\ref{eqn:addsep_hamiltonian_mixedpartial}. The mixed partial derivative of the surrogate, $\hat{f}(x_{1}, y_{1})$, is computed using automatic differentiation via \verb|torch.func| in \verb|pytorch|~\cite{pytorch-autodiff,NEURIPS2019_9015}, at all samples $(x_{1}, y_{1})$ in the test dataset. \verb|torch.func| directly computes the Hessian of the surrogate. For a test dataset comprising $n$ tuples of $(\vec{x}, \vec{y})$, the mixed partial derivative is averaged over all $n$ samples for a comparative evaluation. 

Classifier 8 computes the mixed partial derivative of a surrogate multilayer perceptron neural network symbolically, following Equation~\ref{eqn:method8:deriv}. It models the mixed partial derivative of the unknown function by creating a second surrogate multilayer perceptron neural network with the same architecture. Instead of training the second surrogate multilayer perception neural network, the weights from the surrogate of the unknown function are transferred over. The mixed partial derivative of the surrogate, $\hat{f}(x_{1}, y_{1})$, is computed using the second surrogate multilayer perception neural network at all samples $(x_{1}, y_{1})$ in the test dataset. For a test dataset comprising $n$ tuples of $(\vec{x}, \vec{y})$, the mixed partial derivative is averaged over all $n$ samples for a comparative evaluation.

The activation function of the surrogate multilayer perceptron neural network of the unknown function is the softplus activation function, of the form seen in Equation~\ref{eqn:softplus}. Therefore, the second surrogate multilayer perception neural network that computes the mixed partial derivative also uses the softplus activation function and its derivatives. The first and second derivatives of the softplus activation function are shown in Equations~\ref{eqn:dsoftplus} and \ref{eqn:ddsoftplus} respectively. 
\begin{equation}
    \sigma(x) = \log(\exp(x)+1) \label{eqn:softplus}
\end{equation}
\begin{equation}
    \sigma^\prime_x = \pdv{\sigma({x})}{x} = \frac{\exp(x)}{\exp(x)+1} \label{eqn:dsoftplus}
\end{equation}
\begin{equation}
    \sigma^{\prime\prime}  = \pdv[2]{\sigma({x})}{x} = \frac{\exp(x)}{(\exp(x)+1)^2} \label{eqn:ddsoftplus}
\end{equation}

All code was implemented in \verb|python|. The code to train the surrogates of the unknown functions, and to implement the eight classifiers, is available at \verb|https://github.com/zykhoo/AdditiveSeparabilityTest.git|.


\subsubsection{Evaluation Metrics}
Each classifier computes an average mixed partial derivative for each of the $3744$ unknown functions. A threshold value is set, and if the average mixed partial derivative of an unknown function falls below or is equal to the threshold, the classifier classifies that unknown function as additively separable. Otherwise, the classifier classifies the unknown function as non-additively separable. This is a binary classification problem.

The metric used to compare the eight methods is the accuracy of the classification, using the threshold that gives no false positives as the optimal threshold. The classification accuracy looks at fractions of correctly assigned positive and negative classifications. As half of the unknown functions are additively separable, and half are non-additively separable, the set of functions is balanced. Furthermore, it is equally important to identify both additively separable and non-additively separable unknown functions. Therefore the classification accuracy is used as a metric. The classification accuracy is computed as the count of true positives and negatives, divided by the count of all classifications. A higher accuracy is better. 


\subsection{Experimental Results}
We report the accuracy for the eight classifiers and their respective thresholds in Tables~\ref{tab:accuracy}. Additionally, we report the time taken for each classifier to evaluate a surrogate in Table~\ref{tab:time}

\begin{table}[]
    \centering
    \begin{tabular}{|c|c|c|c|c|c|c|c|c|}
        \hline
        Classifier & 1 & 2 & 3 & 4 & 5 & 6 & 7 & 8 \\
        \hline
        Accuracy & 0.8654 & 0.7647 & 0.7260 & 0.6571 & 0.6343 & 0.6343 & 0.6343 & 0.6343 \\
        Threshold & 0.0109 & 0.0030 & 0.0053 & 0.0018 & 0.0050 & 0.0050 & 0.0050 & 0.0050 \\
        \hline
    \end{tabular}
    \caption{Accuracy and optimal threshold for the eight classifiers. }
    \label{tab:accuracy}
\end{table}

From Table~\ref{tab:accuracy} we observe that generally, all models have high classification accuracy and can be used to classify additively and non-additively separable functions. Classifier 1 performs the best, with an accuracy of 0.8654 at a threshold of 0.0109. Classifiers 2, 3 and 4 also perform well, with accuracies ranging between 0.7647 to 0.6571. Lastly. Classifiers 5, 6, 7 and 8 have an accuracy of 0.6343. It can also be observed from Table~\ref{tab:accuracy} that Classifiers 5, 6, 7 and 8 have the same accuracy and thresholds, as they all compute the mixed partial derivative of the neural network through automatic or symbolic differentiation. These mixed partial derivatives are instantaneous or at the limit, and computed at the same samples in the test dataset and, therefore have the same accuracy. 

Classifiers 1 through 4 outperform Classifiers 5 through 8. Equation~\ref{eqn:addsep_hamiltonian_mixedpartial} implies that for an additively separable function $f(\vec{x},\vec{y}) = g(\vec{x}) + h(\vec{y})$, Equations~\ref{eqn:mixed_distance1} and \ref{eqn:mixed_distance2} hold, as the functions $g(\vec{x})$ and $h(\vec{y})$ are independent of $\vec{y}$ and $\vec{x}$ respectively. Therefore, Equations~\ref{eqn:mixed_distance1} and \ref{eqn:mixed_distance2} should also hold even when the change in $y_n$ or $x_m$ is large, or when $h$ and $k$ is large. Classifiers 1 through 4, when computing the mixed partial derivative of a surrogate via finite difference, check that Equations~\ref{eqn:mixed_distance1} and \ref{eqn:mixed_distance2} hold even for large changes in $y_n$ and $x_m$, therefore outperform Classifiers 5 through 8 that only check to ensure that Equations~\ref{eqn:mixed_distance1} and \ref{eqn:mixed_distance2} hold at the limit. 
\begin{equation}
    \pdv{g(\vec{x})}{y_n} = 0 \label{eqn:mixed_distance1}
\end{equation}
\begin{equation}
    \pdv{h(\vec{y})}{x_m} = 0 \label{eqn:mixed_distance2}
\end{equation}

We make two notes about the observation above. Firstly, Classifiers 5 through 8 compute an instantaneous derivative whose magnitude may depend on the analytical form of the function. A non-additively separable function may have a small instantaneous derivative, and therefore be mistaken as additively separable when using Classifiers 5 through 8. For example, $f(\vec{x},\vec{y}) = xy$ can have a small instantaneous derivative with respect to $x$ of $y$ and may potentially be classified as additively separable. Classifiers 1 to 4, by computing the partial derivative via finite difference instead of instantaneous derivative, compute the change in $f(\vec{x},\vec{y})$ and can check if $f(\vec{x},\vec{y})$ changes greatly when $y$ increases to identify that it is non-additively separable. Secondly, Classifiers 1 and 2 outperform Classifiers 3 and 4 because they compute the finite difference over larger changes in $x_n$ and $y_m$. The former takes the distance between any two samples in the test data, while the latter takes the distance between any sample and the median of the test data. 

It is for this same reason that Classifiers 1 and 3 outperform Classifiers 2 and 4. The latter normalise Classifiers 1 and 3 by the magnitude of $x_n$ and $y_m$. The normalisation obscures the effect of computing the finite differences. 




Lastly, from Table~\ref{tab:time}, we observe that generally, Classifiers 5, 6 and 8 are the most time efficient. Classifiers 1 and 2 are time-consuming because they compute $n^2$ computations. Classifier 7 is also time-consuming because it computes the Hessian of the neural network at each sample in the test dataset. This involves not just computing the mixed partial derivative of the multilayer perceptron neural network, but all other second-order derivatives as well. 
\begin{table}[]
    \centering
    \begin{tabular}{|c|c|c|c|c|c|c|c|c|}
        \hline
        Classifier & 1 & 2 & 3 & 4 & 5 & 6 & 7 & 8 \\
        \hline
        Time &  48.3195 & 52.1720 & 0.0034 & 0.0032 & 0.0025 & 0.0025 & 136.4385 & 0.0029\\
        \hline
    \end{tabular}
    \caption{Time taken in seconds for the eight classifiers to evaluate a surrogate over a test dataset. }
    \label{tab:time}
\end{table}

\section{Conclusion} \label{sec:conclusion}
We presented and comparatively and empirically evaluated the performance of eight classifiers for additive separability, to be used to compute the mixed partial derivatives of surrogate functions. 
Classifier 1 is the most effective, followed by Classifier 2 and Classifier 3. Classifiers 5, 6 and 8 are the most efficient, followed by Classifiers 3 and 4. Classifier 3 is the test of choice given a time constraint. 

The surrogate of a function can be tested for additive separability using the methods introduced in this paper. The detection that the surrogate is additively separable can be leveraged to improve further learning. 
We are now working on the embedding of information regarding additive separability into a multilayer perceptron neural network surrogate that has been detected to be additively separable.

\section*{Acknowledgements}
This research is supported by Singapore Ministry of Education, grant MOE-T2EP50120-0019, and by the National Research Foundation, Prime Minister’s Office, Singapore, under its Campus for Research Excellence and Technological Enterprise (CREATE) programme as part of the programme Descartes.

\bibliographystyle{splncs04}
\bibliography{cite}

\end{document}